%% file: main.tex
\documentclass[10pt,conference]{IEEEtran}
\IEEEoverridecommandlockouts
\usepackage{cite}
\usepackage{amsmath,amssymb,amsfonts}
\usepackage{algorithmic}
\usepackage{graphicx}
\usepackage{textcomp}
\usepackage[font=small, labelformat=simple]{caption}
\usepackage{bm}
\usepackage{xcolor}
\usepackage{booktabs}
\usepackage{multirow}
\usepackage{pifont}
\usepackage{algorithm}
\usepackage{lineno}
\usepackage{wasysym}
\usepackage{enumitem}
\usepackage{colortbl}

\usepackage{xspace}

\def\BibTeX{{\rm B\kern-.05em{\sc i\kern-.025em b}\kern-.08em
    T\kern-.1667em\lower.7ex\hbox{E}\kern-.125emX}}
\begin{document}

\def\model{aiXcoder-7B\xspace}
\def\eg{\textit{e.g.,} }
\def\ie{\textit{i.e.,} }

\title{\model: A Lightweight and Effective Large Language Model for Code Processing}

\author{
\IEEEauthorblockN{Siyuan Jiang$^1$ \thanks{1. Siyuan Jiang and Jia Li contribute equally and are co-first authors.}}
\IEEEauthorblockA{
\textit{aiXcoder} \\
Beijing, China \\
jiangsiyuan@aixcoder.com}
\and
\IEEEauthorblockN{Jia Li$^1$ $\male$}
\IEEEauthorblockA{
\textit{Peking University} \\
Beijing, China \\
lijia@stu.pku.edu.cn}
\and
\IEEEauthorblockN{He Zong}
\IEEEauthorblockA{
\textit{aiXcoder} \\
Beijing, China \\
zonghe@aixcoder.com}
\and
\IEEEauthorblockN{Huanyu Liu, Hao Zhu}
\IEEEauthorblockA{
\textit{Peking University} \\
Beijing, China \\
\{huanyuliu, zhuhao\}@stu.pku.edu.cn}
\and
\IEEEauthorblockN{Shukai Hu, Erlu Li, Jiazheng Ding, Yu Han, Wei Ning, Gen Wang}
\IEEEauthorblockA{
\textit{aiXcoder} \\
Beijing, China \\
\{hushukai, lierlu, dingjiazheng, hanyu, ningwei, wanggen\}@aixcoder.com}
\and
\IEEEauthorblockN{Yihong Dong, Kechi Zhang, Ge Li$^2$ \thanks{2. Corresponding author.}}
\IEEEauthorblockA{
\textit{Peking University} \\
Beijing, China \\
dongyh@stu.pku.edu.cn \\ 
\{zhangkechi, lige\}@pku.edu.cn}
}

\maketitle

\input{chapter/0_Abstract}


\input{chapter/1_Introduction}

\input{chapter/3_Data_Collection}

\input{chapter/4_Training}

\input{chapter/5_Experiments}

\input{chapter/6_Results}

\input{chapter/7_Discussion}

\input{chapter/8_Related_Work}

\input{chapter/9_Conclusion}

\section*{Acknowledgment}

This research is supported by the National Key R\&D Program under Grant No. 2023YFB4503801, the Major Program (JD) of Hubei Province (No.2023BAA024), and the National Natural Science Foundation of China under Grant No. 62192733, 62192730, 62072007.

\bibliographystyle{IEEEtran}
\bibliography{reference}

\end{document}

%% file: chapter/0_Abstract.tex
\begin{abstract}
Large Language Models (LLMs) have been widely used in code completion, and researchers are focusing on scaling up LLMs to improve their accuracy. However, larger LLMs have lower inference efficiency, affecting developers' experience and productivity. In this paper, we propose a lightweight and effective LLM for code completion named \model. Compared to existing LLMs, \model achieves higher code completion accuracy while having smaller scales (\ie 7 billion parameters). We attribute the superiority of \model to three key factors: \ding{182} Multi-objective training. We employ three training objectives, one of which is our proposed Structured Fill-In-the-Middle (SFIM). SFIM considers the syntax structures in code and effectively improves the performance of LLMs for code. 
\ding{183} Diverse data sampling strategies. They consider inter-file relationships and enhance the capability of LLMs in understanding cross-file contexts.
\ding{184} Extensive high-quality data. We establish a rigorous data collection pipeline and consume a total of 1.2
trillion unique tokens for training \model. This vast volume of data enables \model to learn a broad code distribution.
We evaluate \model in five popular code completion benchmarks and a new benchmark collected by this paper. 
The results show that \model outperforms the latest seven LLMs with similar sizes and even surpasses four larger LLMs (\eg StarCoder2-15B and CodeLlama-34B), positioning \model as a lightweight and effective LLM for academia and industry. Finally, we summarize three valuable insights for helping practitioners train the next generations of LLMs for code. \model has been open-sourced and gained significant attention \cite{aiXcoder-7B}. Until January 2025, \model has received 2,226 GitHub Stars.
\end{abstract}

%% file: chapter/1_Introduction.tex
\section{Introduction}
\label{sec:intro}

Large Language Models (LLMs) have been widely used in code completion \cite{CodeLlama,DeepSeek-Coder,StarCoder,StarCoder2}, \ie predicting the subsequent code based on the previous context. For example, GitHub Copilot \cite{Copilot}, an LLM-based code completion tool, is regularly utilized by developers from over 10,000 organizations. Nowadays, researchers often improve the accuracy of LLMs by scaling up LLMs, \eg CodeLlama-70B \cite{CodeLlama}. However, larger LLMs will increase the response time of code completion, which is a critical factor for developer experience and productivity. Thus, it is necessary to train lightweight LLMs that maintain high code completion accuracy while having smaller scales.

Recognizing the above research gap, we present \model, a lightweight and powerful LLM for code completion. \model contains 7 billion parameters, ensuring a high inference speed while achieving superior code completion accuracy. In our later experiments, \textbf{\model outperforms the latest LLMs with similar sizes in six code completion benchmarks and even surpasses larger LLMs (\eg StarCoder2-15B and CodeLlama-34B).} aiXcoder-7B effectively balances model size and performance, providing a better foundational model for both academia and industry.

Compared to previous LLMs, we attribute the superiority of \model to the following three key factors:
\begin{itemize}[leftmargin=*]
    \item \textbf{Multi-objective training.} Previous LLMs mainly use Next-Token Prediction (NTP) as the training objective, which only covers limited code completion scenarios. To address this limitation, \textbf{we propose multi-objective training, including NTP, Fill-In-the-Middle (FIM), and Structured Fill-In-the-Middle (SFIM).} NTP simulates the scenario where developers write a new file from top to bottom, and FIM models the scenario of developers modifying existing code. Because FIM mainly trains models to predict incomplete and irregular code snippets, we further propose SFIM. It parses the code into a syntax tree and mines a relatively complete code span based on a tree node. \model is trained to predict the code span based on its surrounding context. The three objectives help \model learn a comprehensive code completion ability across a wider range of code completion scenarios. Details of the multi-objective training are in Section \ref{sec:training:objective}.
    \item \textbf{A diverse data sampling algorithm.} A code repository often contains multiple code files. Previous studies \cite{CodeLlama,StarCoder,StarCoder2} typically randomly sample files for training, failing to leverage the relationships and contextual information between files. \textbf{We propose four new sampling strategies: sampling based on file content similarity, sampling based on file path similarity, sampling based on inter-file dependency, and random sampling.} The first three strategies simulate common cross-file code completion scenarios, such as code completion augmented by similar code and cross-file API completion, helping \model better understand and utilize dependencies across files. The fourth strategy, random sampling, is to simulate other potential code completion scenarios. Through these diverse sampling strategies, we enhance \model's understanding capability of cross-file contexts within a repository. Details of our data sampling algorithm are in Section \ref{sec:training:data_sampling}.
    \item \textbf{Extensive high-quality data.} LLMs are inherently data-driven, and their performance is significantly influenced by the quantity and quality of the training data. We establish a rigorous data collection pipeline, including data crawling, data cleaning, deduplication, code quality checks, and sensitive information removal. We leverage this pipeline to collect a substantial amount of high-quality training data. \textbf{We continuously feed the training data into \model, consuming a total of 1.2 trillion unique tokens.} This vast volume of data enables \model to learn a broad distribution of code data, allowing it to perform exceptionally well across different code completion scenarios. More details of our data collection pipeline are in Section \ref{sec:data_collection}.
\end{itemize}

We assess the effectiveness of \model in three code completion tasks, including Fill-In-the-Middle (FIM), cross-file code completion, and Natural language to Code (NL2Code). We experiment with six code completion benchmarks, five of which are popular public datasets and one is FIM-Eval collected by this paper. FIM-Eval is a benchmark for FIM, consisting of 16,136 samples and covering four languages (\ie Java, Python, C++, JavaScript). FIM-Eval additionally labels the types of code to be completed, including 13 types, \eg function signatures and comments. Then, we compare \model to 11 recently released LLMs (from 7B to 34B) on these benchmarks and yield the following insights: 
\ding{182} \model substantially outperforms LLMs with similar sizes and even surpasses larger LLMs in six benchmarks. For example, in a popular benchmark - HumanEval, \model achieves a Pass@1 score of 54.9\%, outperforming CodeLlama-34B (\ie 48.2\%) and StarCoder2-15B (\ie 46.3\%). The improvements show that \model achieves higher code completion accuracy while having smaller scales.
\ding{183} Based on our FIM-Eval, we analyze the performance of \model in completing different types of code. \model outperforms LLMs with similar sizes on most types (max: 13, min: 8). The results show the strong generalization ability of \model in code completion.
\ding{184} We show that existing LLMs are prone to generate longer code in FIM, while the code generated by \model is closer in length to human-written reference code. The result shows that the code generated by \model is more concise and closer to the human coding style.

\textbf{Insights of training LLMs for code.} Based on our practices in \model, we summarize three valuable insights, including scaling up training data and introducing the inter-file relationships and code structures into the training. These insights can help practitioners train the next generations of LLMs for code.

We summarize the key contributions of this paper:
\begin{itemize}[leftmargin=*]
    \item We present \model, a lightweight and effective LLM with 7 billion parameters for code completion. We have released its weights and code \cite{aiXcoder-7B}. As of the submission date, \model has received 2,193 GitHub Stars.
    \item We propose a novel training objective - Structured Fill-In-the-Middle, which considers the syntax structures in code and effectively improves the performance of LLMs.
    \item We propose a new data sampling algorithm for code, which considers inter-file relationships and enhances the capability of LLMs in understanding cross-file contexts.
    \item We release a new code completion benchmark, consisting of 16,136 samples and covering four languages.
    \item We evaluate the effectiveness of \model in six code completion benchmarks. \model substantially outperforms 7 LLMs with similar sizes and even surpasses 4 larger LLMs (15B and 34B).
\end{itemize}

\textbf{Paper Organization.} Section \ref{sec:data_collection} presents our data collection pipeline. Section \ref{sec:training} describes the training process of \model. Section \ref{sec:study_design} and Section \ref{sec:results} show the study design and results, respectively. Section \ref{sec:discussion} discusses the performance of \model in other aspects and potential threats. Section \ref{sec:related_work} summarizes the recently proposed LLMs. Finally, Section \ref{sec:conclusion} concludes this paper and points out future directions.

%% file: chapter/3_Data_Collection.tex
\begin{figure*}[htbp]
\centering
\includegraphics[width=0.99\textwidth]{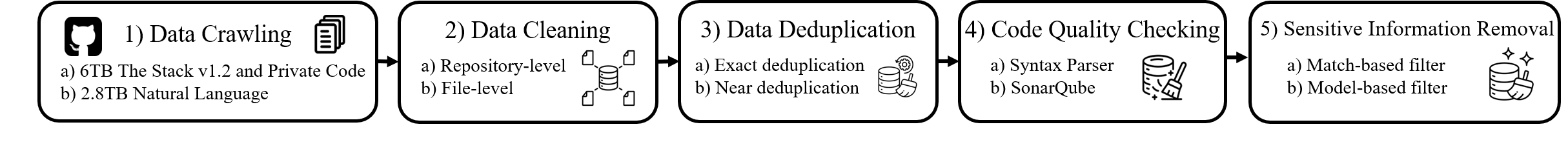}
\vspace{-0.2cm}
\caption{An overview of our data collection pipeline.}
\label{data_overview}
\end{figure*}

\section{Data Collection Pipeline}
\label{sec:data_collection}

\begin{figure}[htbp]
\centering
\includegraphics[width=0.9\linewidth]{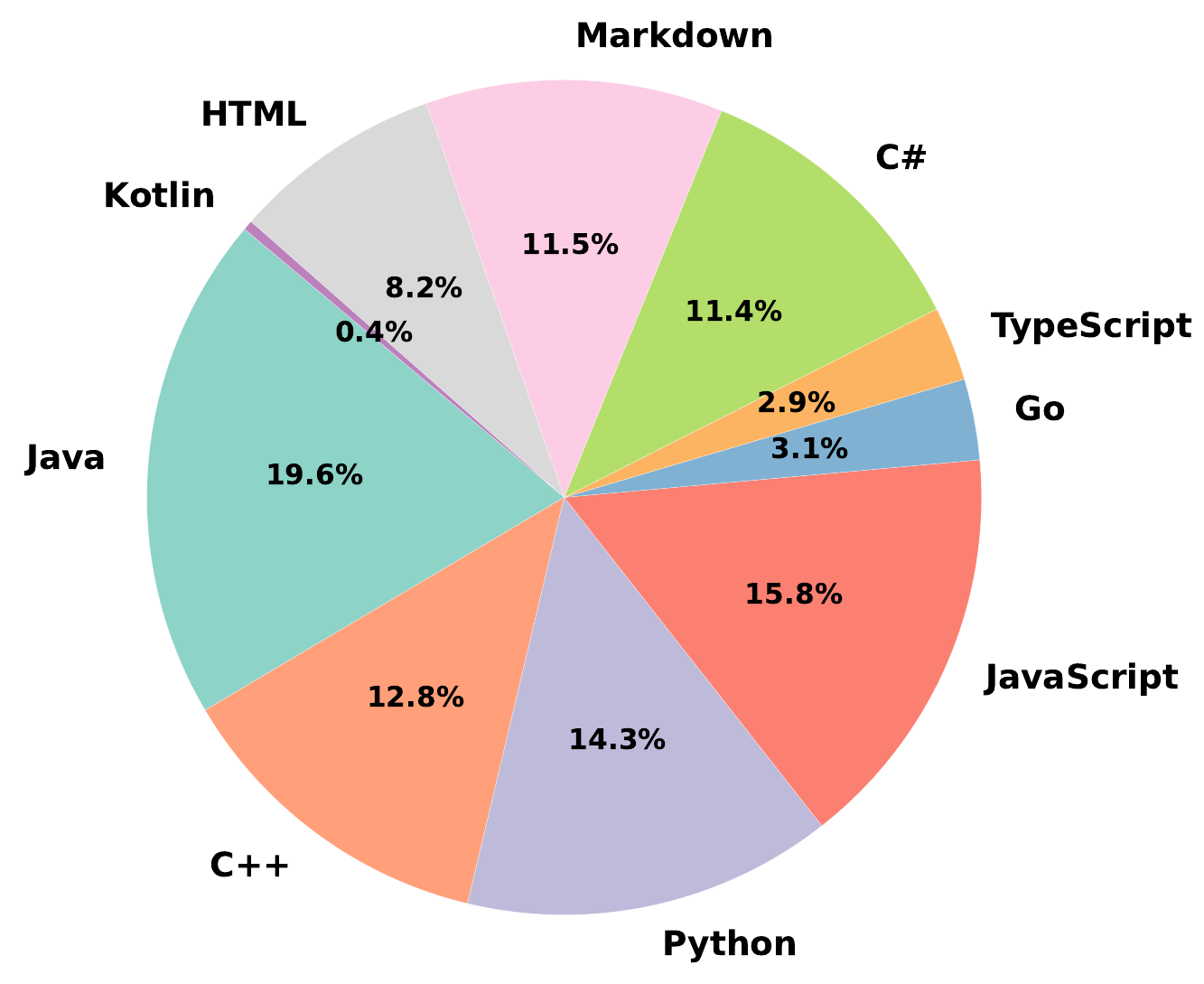}
\caption{The distributions of the top 10 programming languages in our source code training data.}
\label{fig:language_distribution}
\end{figure}

This section presents the process of collecting the pre-training data of \model. Figure \ref{data_overview} shows an overview of our data collection pipeline, consisting of five stages: data crawl (Section \ref{sec:data_collection:data_crawl}), data cleaning (Section \ref{sec:data_collection:data_cleaning}), data deduplication (Section \ref{sec:data_collection:data_dedup}), code quality checking (\ref{sec:data_collection:code_analysis}), and sensitive and personally identifiable information removal (Section \ref{sec:data_collection:PII}). Through this pipeline, we collect and clean 2.8TB of natural language data and 3.5TB of source code data. Figure \ref{fig:language_distribution} visualizes the distributions of the top 10 programming languages in the code data. Next, we describe the details of the data collection pipeline in the following sections. 

\subsection{Data Crawling}
\label{sec:data_collection:data_crawl}

The pre-training data of \model consists of two parts: natural language data and source code data. 

\textbf{Natural Language Data.} We collect natural language data from two public datasets: WuDaoCorpora \cite{WuDaoCorporaText} and RefineWeb \cite{RefinedWeb}, driven by two key motivations. First, these datasets are highly diverse, covering a wide range of domains and languages. They include a broad spectrum of natural language text from the internet, such as social media conversations, books, and technical papers, and cover two mainstream languages, \ie English and Chinese. Second, both datasets have been thoroughly cleaned and deduplicated in previous studies, which significantly reduces the preprocessing workload and allows us to focus on processing code data. Finally, we collect 2.8TB of natural language data for pre-training.

\textbf{Source Code Data.} The raw source code data comes from two sources: one is the open-source dataset - The Stack v1.2, and the other is the code data we crawled ourselves. 

\begin{itemize}[leftmargin=*]
    
    \item \textit{The Stack v1.2} \cite{TheStackDatasets} is a comprehensive dataset comprising approximately 6TB of permissively licensed source code sourced from public GitHub repositories, spanning 358 programming languages, with notable representation from HTML, JavaScript, Java, and C. This dataset has undergone rigorous cleaning processes to enhance data integrity and prevent inflated performance metrics due to duplicate content. Only repositories with permissive licenses were retained, while non-essential files, such as binaries and those exceeding 1MB, were systematically excluded. The version 1.2 of The Stack has excluded opt-out requests submitted by February 9, 2023, as well as initially flagged malicious files (this exclusion is not exhaustive).
    
    \item \textit{Our crawled data.} Popular repositories we have been crawling from GitHub for the past decade.
\end{itemize}

\subsection{Data Cleaning}
\label{sec:data_collection:data_cleaning}

In this stage, we clean the collected data by removing invalid or low-quality data. Because the natural language data has already undergone rigorous cleaning, we focus on cleaning the source code data. Our cleaning process comprises two steps: repository-level cleaning and file-level cleaning. Below, we provide a detailed explanation of each step.

\textbf{Repository-level Cleaning.} Our goal is to remove repositories with impressive licenses and low-quality repositories. To achieve this goal, our cleaning is performed in three steps:
\begin{itemize}[leftmargin=*]
    \item \textit{Collecting permissive licenses.} We build a list of permissive licenses based on the Blue Oak Council\footnote{https://blueoakcouncil.org/list} and previous work \cite{TheStackDatasets}. This list includes various permissive licenses with minimal restrictions on software copying, modification, and redistribution. Only repositories with licenses from this list are retained for pre-training.
    \item \textit{Identifying repositories' licenses.} GHArchive provides license information when repository owners explicitly set the code license through the web interface. We first extract each repository's license from GHArchive. If a license is not listed in GHArchive, we leverage the go-license-detector\footnote{https://github.com/src-d/go-license-detector} to identify the most likely license.
    \item \textit{Removing repositories with impressive licenses.} After identifying the licenses, we exclude repositories whose licenses do not appear on the permissive license list.
    \item \textit{Removing low-quality repositories.} We score the repositories from different aspects, including the number of stars, the number of git commits, and the number of test files. Then, we sort the repositories in descending order based on their scores and remove the lowest 10\%.
\end{itemize}

\textbf{File-level Cleaning.} Next, we filter out low-quality files in repositories. Specifically, we empirically design some rules to filter out low-quality documents: \ding{182} Trivial files, including empty files, corrupted files, non-text files, and auto-generated files. \ding{183} Too long files. Too long files typically contain wordy or repetitive content and are not suitable as training data. If a line in a file exceeds 1000 characters, the total number of lines in the file exceeds 10,000, or the size of the file exceeds 1MB, we consider it a long file.

\subsection{Data Deduplication}
\label{sec:data_collection:data_dedup}

Previous work \cite{RefinedWeb} has shown that data deduplication can significantly improve the performance of trained models. It is particularly necessary for code data, where code reuse leads to a large amount of duplicate content. Therefore, in this stage, we eliminate duplicate code files within the repositories. Our deduplication process consists of two steps:
\begin{itemize}[leftmargin=*]
    \item \textit{Exact deduplication.} We extract file contents, find the files with the same content, and keep only one copy.
    \item \textit{Near deduplication.} Exact deduplication is too strict and may cause false positives. Thus, we further perform near deduplication. We compute the MinHash \cite{MiniHash} with 256 permutations of all files and use Locality Sensitive Hashing \cite{LSH} to find clusters of duplicates. We further reduce the clusters by ensuring that each file in the original cluster is similar to at least one other file in the reduced cluster. We consider two files near-duplicate when their Jaccard similarity exceeds 0.85.
\end{itemize}

\subsection{Code Quality Checking}
\label{sec:data_collection:code_analysis}

In this stage, we use code analysis tools to assess the quality of code data and filter out low-quality code. Low-quality code often contains syntax errors, code defects, vulnerabilities, and misleading models that generate unreliable code. Specifically, we use the following tools to assess the quality of code:

\textbf{Syntax Parser.} Syntax correctness is one of the basic principles that source code should satisfy. We use a public syntax parser - tree-sitter\footnote{https://tree-sitter.github.io/tree-sitter/} to parse all code files and delete files that fail to parse or time out.

\textbf{SonarQube.} SonarQube\footnote{https://www.sonarsource.com/products/sonarqube/} is an open-source tool for the inspection of code quality. It can detect code defects, vulnerabilities, code smells, and technical debt in various programming languages. We use SonarQube to identify problematic code files and delete them.

\subsection{Sensitive Information Removal}
\label{sec:data_collection:PII}

In this section, we remove the sensitive information in the pre-training data, \eg texts involving sensitive topics and personally identifiable information (PII). We remove this information in two steps:

\textbf{Match-based filter.} We manually build a list of sensitive words, which covers a broad range of sensitive topics (\eg politics). Then, we scan all pre-training data and delete the data containing sensitive words. For example, if the frequency of sensitive words in a repository of a user exceeds a certain threshold, all repositories of the user will be deleted.

\textbf{Model-based filter.} Following previous work \cite{StarCoder}, we use a Named Entity Recognition (NER) model to identify PII in the data. Specifically, we reuse a trained NER model in previous work \cite{StarCoder}, which can identify six categories of PII, including emails, names, IP addresses, usernames, passwords, and keys. Then, we replace the detected PII entities with the following special tokens: \texttt{<EMAIL>}, \texttt{<NAME>}, \texttt{<IP\_ADDRESS>}, \texttt{<USERNAME>}, \texttt{<PASSWORD>}, \texttt{<KEY>}.

%% file: chapter/4_Training.tex
\section{Model Training}
\label{sec:training}

In this section, we describe the pre-training procedure of \model, including model architecture, data sampling algorithm, and training objectives.

\subsection{Model Architecture}
\label{sec:training:architecture}

\model is built upon an auto-regressive dense Transformer architecture\cite{Transformer}. \model consists of 32 Transformer decoder layers, with a hidden state size of 4096 and an intermediate size of 14464. More details are in our open-sourced repository \cite{aiXcoder-7B}. Our tokenizer is trained with SentencePiece \cite{BPE} upon 500GB of training data. The vocabulary size is 49,512. We adopt Rotary Positional Encodings (RoPE) \cite{RoPE} to enhance the representation of positional information in sequences, following \cite{StarCoder, CodeLlama}. RoPE allows for a more flexible encoding of position than absolute position encoding. Additionally, we implement Grouped Query Attention (GQA) \cite{GQA}, which enhances the efficiency of the attention mechanism by grouping queries, allowing for a more scalable attention computation. We maintain a balanced design in our attention heads, with a configuration of 32 query attention heads and 8 key-value attention heads.

\subsection{Data Sampling Algorithm}
\label{sec:training:data_sampling}

Through the pipeline in Section \ref{sec:data_collection}, we collect extensive code repositories and natural language articles. We randomly shuffle these repositories and articles and iterate through them. If a natural language article is sampled, we process it into training sequences based on the training objectives (Section \ref{sec:training:objective}).

\input{table/sampling_algorithm}

If a code repository is sampled, we design an algorithm for sampling files from the repository, as described in Algorithm \ref{alg:sorting_files}. The algorithm contains four strategies: sampling based on file content similarity, sampling based on file path similarity, sampling based on inter-file dependencies, and random sampling. The first three strategies simulate common cross-file code completion scenarios, such as code completion augmented by similar code and cross-file API completion, helping \model better understand and utilize dependencies across files. The fourth strategy, random sampling, is to simulate other potential code completion scenarios. For each repository, the probability of selecting each of the first three strategies is 30\%, and the probability of selecting the last strategy is 10\%. These sampled files are further converted into training sequences based on the training objectives (Section \ref{sec:training:objective}).

\subsection{Training Objectives}
\label{sec:training:objective}

The training objectives of \model consist of the Next-Token Prediction (NTP) and Structured Fill-In-the-Middle (SFIM), detailed as follows.

\textbf{Next-Token Prediction (NTP).} It is similar to code completion, training models to predict the subsequent token based on the provided context. Given a code file or natural language article $\mathbf{x}=\{x_0, x_1, \dots, x_l\}$, NTP trains models to predict the next token $x_i$ based on previous tokens $\{x_{<i}\}$. The objective is to minimize the following loss function:
\begin{equation}
\operatorname{loss}_{NTP}=-\sum_{i=0}^{l-1} \log p\left(x_i \mid x_{t<i}\right) 
\end{equation}

\textbf{Fill-In-the-Middle (FIM) \cite{FIM}.} The motivation behind this training objective is that human developers frequently modify existing code, \eg inserting new code snippets. Thus, FIM trains models to predict the middle content based on the preceding and following context. Specifically, given a code file or natural language article $\mathbf{x}=\{x_0, \dots, x_l\}$, we randomly select a span of contiguous tokens as the $\mathbf{middle}=\{x_i, \dots, x_j\}$, using the content above the span as the $\mathbf{prefix}=\{x_0, \dots, x_{i-1}\}$ and the content below as the $\mathbf{suffix}=\{x_{j+1}, \dots, x_l\}$. We employ two distinct modes to construct the training sequence: PSM (\ie $[\mathbf{prefix};\mathbf{suffix};\mathbf{middle}]$) or SPM (\ie $[\mathbf{suffix};\mathbf{prefix};\mathbf{middle}]$). $[;]$ means the concatenation of multiple strings using special tokens. Previous work \cite{DeepSeek-Coder} has found that models work best when PSM and SPM account for 50\% each. Thus, we choose the probability of each mode being 50\%. 

Finally, we feed the training sequence into \model and minimize the following loss function:
\begin{equation}
\begin{aligned}
    \operatorname{loss}_{FIM}= & -\log p\left([\mathbf{prefix};\mathbf{suffix};\mathbf{middle}]\right) \\
    & - \log p\left([\mathbf{suffix};\mathbf{prefix};\mathbf{middle}]\right)
\end{aligned}
\end{equation} 

\begin{figure}[t]
\centering
\includegraphics[width=\linewidth]{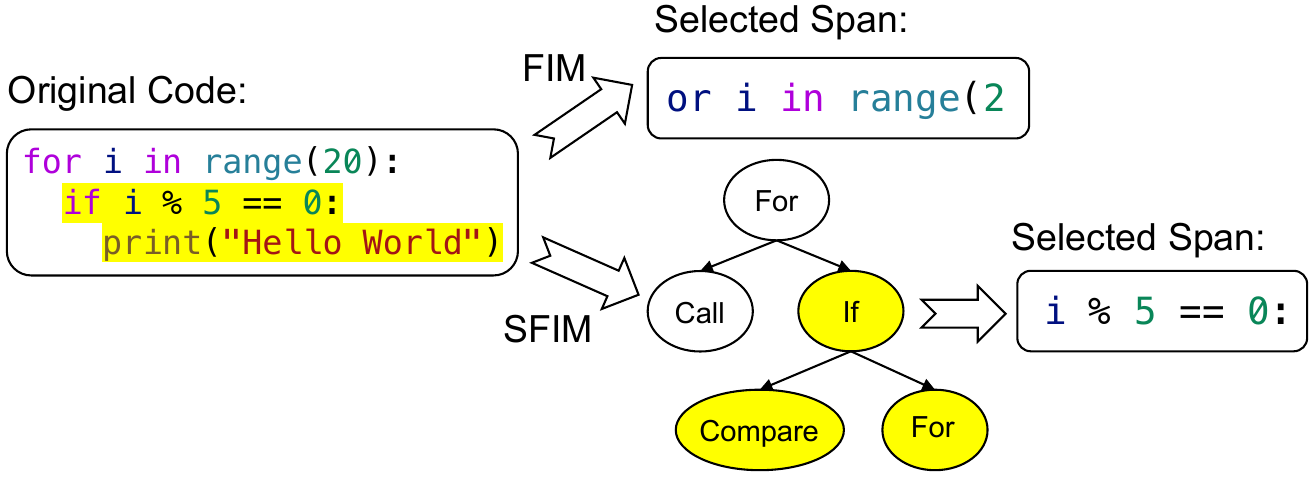}
\caption{Examples of selected spans in FIM and SFIM.}
\label{fig:training_objective}
\end{figure}

\textbf{Structured Fill-In-the-Middle (SFIM).}
As shown in Figure \ref{fig:training_objective}, FIM randomly selects spans and trains models to predict incomplete and irregular code snippets (\eg \texttt{or i in range(2}). However, developers often expect models to complete the current code into a complete snippet, such as a completed code line or loop block, instead of suggesting an incomplete code snippet. To address this, we propose SFIM, which trains \model to predict complete code snippets, enabling \model to align with the practical needs of developers. Given a code file, SFIM uses a new strategy for selecting spans: \ding{182} randomly select a function from the file and parse the function into a syntax tree; \ding{183} randomly choose a non-root, non-leaf node from the tree and locate the code snippet corresponding to this node; \ding{184} within this code snippet, randomly select a span, where the start position is randomly determined, but the end position must be the end of a code line. As shown in Figure \ref{fig:training_objective}, SFIM selects an If node and mines a relatively complete code snippet (\ie \texttt{i\%5==0:}) as the span.

Subsequently, we follow the FIM and convert the select span into a training sequence in the format of PSM or SPM. Based on preliminary experiments, we set the probability of selecting PSM to 30\% and SPM to 70\%. We input the training sequence into \model and minimize the following loss function:
\begin{equation}
\begin{aligned}
    \operatorname{loss}_{SFIM}= & -\log p\left([\mathbf{prefix};\mathbf{suffix};\mathbf{middle}]\right) \\
    & - \log p\left([\mathbf{suffix};\mathbf{prefix};\mathbf{middle}]\right)
\end{aligned}
\end{equation}  

\textbf{Multi-objective Training.} We optimize the above three objectives alternately. Given a code repository, we choose SFIM with a probability of 70\% and FIM and NTP with a probability of 15\%, respectively. Given a natural language article, we choose FIM and NTP with a probability of 50\%, respectively. We determine these probabilities based on previous work \cite{DeepSeek-Coder,StarCoder,StarCoder2} and our preliminary experiments.

\subsection{Training Details}
\label{sec:training:details}

\input{table/training_details}

We leverage Megatron to train \model. The training process is conducted on 128 A100 40GB GPUs, consuming a total of 1.2 trillion unique tokens. The hyper-parameters used during training are shown in Table \ref{table:training_details}.

%% file: table/sampling_algorithm.tex
\begin{algorithm}[htp]
\caption{Sampling code files within a repository.}
\label{alg:sorting_files}
\begin{algorithmic}[1]
    \REQUIRE~~\\ 
        A list of code files $Files$
    \ENSURE~~\\ 
        An ordered list of code files $orderedFiles$
    \STATE $orderedFiles \gets []$
    \STATE $randomValue \gets random(0,1)$
    \IF{$randomValue < 0.3$}
        \STATE \textit{\textcolor{blue}{// Sampling based on file content similarity}}
        \STATE $tfIdfList \gets \text{[]}$
        \FOR{$file$ in $Files$}
            \STATE $tfIdfList.append(\operatorname{TFIDF}(file))$
        \ENDFOR
        \STATE $clusterNum \gets min((random(1, 20), Files.size))$
        \STATE $Clusters \gets \operatorname{KMEANS}(clusterNum, tfIdfList)$
        \FOR{$Cluster$ in $\operatorname{SHUFFLE}(Clusters)$}
            \STATE $orderedFiles.extend(\operatorname{SHUFFLE}(Cluster))$
        \ENDFOR 
    \ELSIF{$randomValue < 0.6$}
        \STATE \textit{\textcolor{blue}{// Sampling based on file path similarity}}
        \WHILE{$Files.size > 0$}
            \STATE $K \gets random(1, Files.size)$
            \STATE $randomFile \gets random(Files)$
            \FOR{$File$ in $\operatorname{KNN\_PATH}(randomFile, K)$}
                \STATE $orderedFiles.append(File)$
                \STATE $Files.remove(File)$
            \ENDFOR
            \STATE $orderedFiles.append(randomFile)$
        \ENDWHILE
    \ELSIF{$randomValue < 0.9$}
        \STATE \textit{\textcolor{blue}{// Sampling based on file dependencies}}
        \STATE $callGraph \gets \operatorname{CALL\_GRAPH}(Files)$
        \STATE $leafNodes \gets \operatorname{GET\_LEAFS}(callGraph)$
        \WHILE{$leafNodes.isNotEmpty()$}
            \FOR{$node$ in $leafNodes$}
                \STATE $lastPredecessors \gets \{node\}$
                \WHILE{$lastPredecessors.isNotEmpty()$}
                    \FOR{$node$ in $\operatorname{SHUFFLE}(lastPredecessors)$}
                        \FOR{$p$ in $node.predecessors$}
                            \STATE $p.successors.remove(node)$
                            \IF{$p.successors.isEmpty()$}
                                \STATE $leafNodes.add(p)$
                            \ENDIF
                        \ENDFOR
                        \STATE $orderedFiles.append(node.file)$
                    \ENDFOR
                    \STATE $lastPredecessors \gets node.predecessors$
                \ENDWHILE
            \ENDFOR
        \ENDWHILE
    \ELSE
        \STATE \textit{\textcolor{blue}{// Random sampling}}
        \STATE $orderedFiles.extend(\operatorname{SHUFFLE}(Files))$
    \ENDIF
    \RETURN $orderedFiles$
\end{algorithmic}
\end{algorithm}

%% file: table/training_details.tex
\setlength{\tabcolsep}{12pt} 
\begin{table}[h]
\centering
\caption{Training Hyperparameters for \model}
\resizebox{0.8\linewidth}{!}{
\begin{tabular}{l|c}
\toprule
\textbf{Hyperparameter}   & \textbf{\model}  \\ 
\midrule
lr-decay-iters            & 320000                \\
weight-decay              & 0.01                  \\
lr-decay-style            & cosine                \\
clip-grad                 & 1.0                   \\
hidden-dropout            & 0.1                   \\
attention-dropout         & 0.05                  \\
adam-beta1, beat2         & 0.9, 0.98              \\
Batch Size                & 512                   \\
Max Learning Rate         & 1e-5                  \\
Context window       & 32768                 \\
\bottomrule
\end{tabular}
}
\label{table:training_details}
\end{table}

%% file: chapter/5_Experiments.tex
\section{Study Design}
\label{sec:study_design}

We design a large-scale study to evaluate the effectiveness of \model. This section presents the details of our study, including research questions, benchmarks, compared baselines, and evaluation metrics.

\subsection{Research Questions}
\label{sec:study_design:RQs}

Our study aims to answer the following Research Questions (RQs). They evaluate \model in three code completion tasks, including Natural Language to Code (NL2Code), Fill-In-the-Middle (FIM), and cross-file code completion.

\textbf{RQ1: How does \model perform on NL2Code task compared to existing LLMs?} 
NL2Code is the task of completing the source code based on a natural language description or function signature. 

\textbf{RQ2: How does \model perform on Fill-In-the-Middle task compared to existing LLMs?} FIM simulates scenarios where developers modify existing code by predicting the missing middle portion using bidirectional contexts.

\textbf{RQ3: How does \model perform on Cross-File Code Completion compared to existing LLMs?} This task requires completing code by using relevant context from other files within the current repository. 

In the RQs, we apply \model on 6 benchmarks in total. These benchmarks cover 6 programming languages. To show the superiority of \model, we also select 10 popular LLMs as baselines for comparison. Then, we report the execution-based and text-based metrics (Section \ref{sec:study_design:metrics}) of completed programs.
\input{table/code_gen_results}
\subsection{Compared Models}
\label{sec:study_design:baselines}

We select 10 popular LLMs for comparison, and they can be divided into two groups:

\ding{182} \textbf{LLMs with similar sizes.} The first group contains six popular LLMs, which have similar sizes to \model.
\begin{itemize}[leftmargin=*]
    \item \textbf{CodeGen2.5-7B} \cite{CodeGen2}, released by Salesforce, is a 7B parameter model specialized in code generation and understanding, trained on a diverse set of programming languages.
    \item \textbf{CodeGeex2-7B} \cite{CodeGeeX}, developed by Zhipu AI, is a 7B parameter model designed for code completion and bug fixing, leveraging a large corpus of code data.
    \item \textbf{CodeLlama-7B} \cite{CodeLlama}, an open-source model by Meta AI, is a 7B parameter architecture fine-tuned on a vast collection of code and natural language data based on Llama2\cite{Llama2}.
    \item \textbf{CodeShell-7B} \cite{CodeShell}, introduced by Shell AI, is a 7B parameter model focused on shell scripting and code interpretation, trained on a mixture of code and command-line data.
    \item \textbf{StarCoder2-7B} \cite{StarCoder}, from BigCode, is a 7B parameter model trained on The Stack v2 dataset, specializing in code understanding and generation across multiple programming languages.
    \item \textbf{DeepSeekCoder-7B} \cite{DeepSeek-Coder}, by DeepSeek AI, is a 7B parameter model trained on a blend of code and natural language data designed for programming tasks.
    \item \textbf{Qwen2.5-Coder-7B} \cite{Qwen2.5-Coder} is a large language model with 7 billion parameters. It builds on the strong Qwen2.5 \cite{Qwen2} and continues training on a larger scale of code data, including source code, text-code grounding data, and synthetic data, totaling 5.5 trillion tokens. This leads to significant improvements in code-related tasks.
\end{itemize}

\ding{183} \textbf{LLMs with larger sizes.} We also select four larger LLMs as baselines to demonstrate the superiority of \model:
\begin{itemize}[leftmargin=*]
    \item \textbf{CodeLlama-13B}\cite{CodeLlama} is an enhanced version of the CodeLlama model with 13B parameters.
    \item \textbf{StarCoder-15B}\cite{StarCoder} is the expanded version of the StarCoder model with 15B parameters, delivering improved accuracy for code synthesis and interpretation.
    \item \textbf{StarCoder2-15B}\cite{StarCoder2} is a 15B parameter model upgraded on the original StarCoder, offering refined code generation and more diverse programming languages.
    \item \textbf{CodeLlama-34B}\cite{CodeLlama} is the largest variant of the CodeLlama series with 34B parameters.
\end{itemize}

\subsection{Benchmarks}
\label{sec:study_design:benchmarks}

\textbf{NL2Code Benchmarks.} Following previous studies \cite{DeepSeek-Coder,AceCoder,SCoT}, we select three popular NL2Code benchmarks in our experiments, detailed as follows.

\begin{itemize}[leftmargin=*]
    \item \textbf{HumanEval \cite{HumanEval} and MBPP \cite{MBPP}} consist of 164 and 974 Python programming problems. Each problem includes a function signature, a detailed docstring, and several test cases. LLMs are required to complete the function body based on the signature and docstring. The generated code is checked by executing test cases, being considered correct only if all tests pass.
    \item \textbf{MultiPL-E \cite{MultiPL-E}} is the multilingual version of HumanEval, covering multiple programming languages, \eg C++, Java, and JavaScript. 
\end{itemize}

\textbf{FIM Benchmarks.} Code is rarely composed in a straightforward left-to-right sequence. Simulating when a developer modifies existing code, FIM refers to the task of completing missing a middle code snippet leveraging bidirectional contexts.

\begin{itemize}[leftmargin=*]
    \item \textbf{Santacoder-data \cite{Santacoder}} is a popular FIM benchmarks consisting of 4,792 samples. It is built from MultiPL-E \cite{MultiPL-E} and requires LLMs to predict a single line of code based on the preceding and following context.

    \item \textbf{FIM-Eval} is a large-scale FIM benchmark collected by this paper. We construct FIM-Eval from some real-world repositories, which are excluded from the training data of \model. We extract 13 types of code snippets from these repositories and randomly mine spans from these code snippets.
    These 13 types of code snippets encompass common code completion scenarios, including method signatures, method bodies, single-line statements, methods with comments, empty code blocks, specific positions within a method body (top, middle, and bottom), specific control statements (\ie if statements, for loops, while loops, try statements, and switch-case statements).
    Finally, we collected 16,140 samples covering four programming languages: C++ (4,080 samples), Java (4,080 samples), Python (3,900 samples), and JavaScript (4,080 samples). FIM-Eval provides a reliable, practical, and diverse evaluation platform for FIM. FIM-Eval has been open-sourced in our repository \cite{aiXcoder-7B}.
\end{itemize}

\textbf{Cross-File Code Completion Benchmarks.} This task requires LLMs to complete the code based on cross-file context within the same project. Building upon insights from prior research \cite{DeepSeek-Coder, StarCoder}, detailed as follows.

\begin{itemize}[leftmargin=*]
    \item \textbf{CrossCodeEval \cite{CrossCodeEval}} covers four popular programming languages: 2,665 Python samples, 2,139 Java samples, 3,356 TypeScript samples, and 1,768 C\# samples. Each sample is provided in three formats: no cross-file context, retrieved cross-file context, and retrieval with reference. The completed code snippet is compared using text-based metrics.
\end{itemize}

\subsection{Evaluation Metrics}
\label{sec:study_design:metrics}

We describe the evaluation metrics used in different code completion tasks.

\textbf{NL2Code}. NL2Code benchmarks provide test cases for evaluation. Thus, we execute test cases to check the correctness of the generated code and report Pass@$k$ \cite{HumanEval}. Specifically, we generate $n \geq k$ code snippets per testing sample, count the number of correct code snippets $c \leq n$ that pass all test cases, and calculate the Pass@$k$:
\begin{equation}
\text{Pass}@k:=\underset{\text { Samples }}{\mathbb{E}}\left[1-\frac{\left(\begin{array}{c}
n-c \\
k
\end{array}\right)}{\left(\begin{array}{l}
n \\
k
\end{array}\right)}\right]
\end{equation}

\textbf{FIM and cross-file code completion.} We consider the LLMs' completions as predictions and the human-written completions as references. We compare the predictions to references and compute the following metric:
\begin{itemize}[leftmargin=*]
    \item \textbf{BLEU \cite{Bleu}} measures the $n$-gram similarity between predictions and references. $n$ is empirically set to 4.
    \item \textbf{CodeBLEU \cite{CodeBleu}} is a variant of BLEU for code. It considers not only the $n$-gram similarity but also the syntax and data flow similarity. 
    \item \textbf{Exact Match (EM)} evaluates the percentage of cases where the prediction exactly matches the reference, providing a strict measure of how often LLMs produce correct code without deviations.
    \item \textbf{Edit Similarity (ES)} measures the similarity between the prediction and the reference based on the number of edits required to transform one into the other, typically using metrics like Levenshtein distance \cite{levenshtein1966binary}.
\end{itemize}

%% file: table/code_gen_results.tex
\begin{table*}[t]
    \centering
    \caption{The Pass@1 of LLMs on NL2Code benchmarks.}
    \label{table:code_gen_results}
    \resizebox{0.95\linewidth}{!}{
    \begin{tabular}{lcccccc}
        \toprule
        Model & HumanEval & MBPP & MultiPL-E (C++) & MultiPL-E (Java) & MultiPL-E (JS) & Average \\
           \midrule
            CodeGen2.5-7B & 28.7\% & 39.2\% & 25.7\% & 26.1\% & 26.2\% & 29.1\% \\
            CodeGeex2-7B & 36.0\% & 36.2\% & 29.2\% & 25.9\% & 24.8\% & 30.4\% \\
            CodeLlama-7B & 31.7\% & 38.6\% & 29.8\% & 34.2\% & 29.2\% & 32.7\% \\
            CodeShell-7B & 34.4\% & 38.6\% & 28.2\% & 30.4\% & 33.2\% & 32.9\% \\
            StarCoder2-7B & 35.4\% & 54.4\% & 33.6\% & 29.4\% & 35.4\% & 37.6\% \\
            DeepSeekCoder-7B & 49.4\% & 60.6\% & 50.3\% & 43.0\% & 48.4\% & 50.3\% \\
            \rowcolor[rgb]{ .741,  .843,  .933}
            \model & \textbf{54.9\%} & \textbf{66.0\%} & \textbf{58.2\%} & \textbf{57.0\%} & \textbf{64.5\%} & \textbf{60.1\%} \\
            \midrule
            StarCoder-15B & 31.7\% & 42.8\% & 31.1\% & 28.5\% & 29.8\% & 32.8\% \\
            CodeLlama-13B & 36.0\% & 48.4\% & 37.9\% & 38.0\% & 32.3\% & 38.5\% \\
            StarCoder2-15B & 46.3\% & 66.2\% & 41.4\% & 33.9\% & 44.2\% & 46.4\% \\
            CodeLlama-34B & 48.2\% & 55.2\% & 44.7\% & 44.9\% & 42.2\% & 47.0\% \\
        \bottomrule
        \end{tabular}}
\end{table*}

%% file: chapter/6_Results.tex
\section{Results and Analyses}
\label{sec:results}

\input{table/santacoder_overall}

\subsection{RQ1: Performance on NL2Code}
\label{sec:results:nl2code}

Following recent work on LLMs \cite{DeepSeek-Coder, StarCoder2}, we use greedy decoding and report Pass@$1$. Table \ref{table:code_gen_results} shows the results of different LLMs on NL2Code benchmarks. From Table \ref{table:code_gen_results}, we draw the following observations:
\begin{itemize}[leftmargin=*]
    \item Compared to LLMs of similar sizes, our \model achieves the current best results, outperforming the top-performing model DeepSeekCoder-7B by an average of 9.8\%. Moreover, it significantly surpasses CodeGen2.5-7B with a 31\% absolute advantage.
    
    \item \model even surpasses four larger LLMs (\eg StarCoder2-15B and CodeLlama-34B), achieving a lead of 13.1\% over CodeLlama-34B, which is nearly five times larger, and 13.7\% over StarCoder2-15B on average.
    
    \item Across languages like Java, Python, C++, and JavaScript, our aiXcoder-7B shows strong performance. It surpasses DeepSeekCoder-7B by 16.1\% in JavaScript and exceeds by 5.5\% in Python.
\end{itemize}
\input{table/code_gen_cross}
\input{table/extend_data}

\subsection{RQ2: Performance on Fill-In-the-Middle (FIM)}
\label{sec:results:code_infilling}

Generally, FIM closely mirrors how developers modify existing code, making it an ideal method for evaluating models in real-world programming scenarios. 

Based on the experimental findings outlined in Table \ref{table:result-santacoder}, \model demonstrates the highest overall performance on \textbf{SantaCoder-data}, achieving the best results in Python, JavaScript, and Java among the models tested.

\begin{figure*}[ht]
\centering
\includegraphics[width=0.99\textwidth]{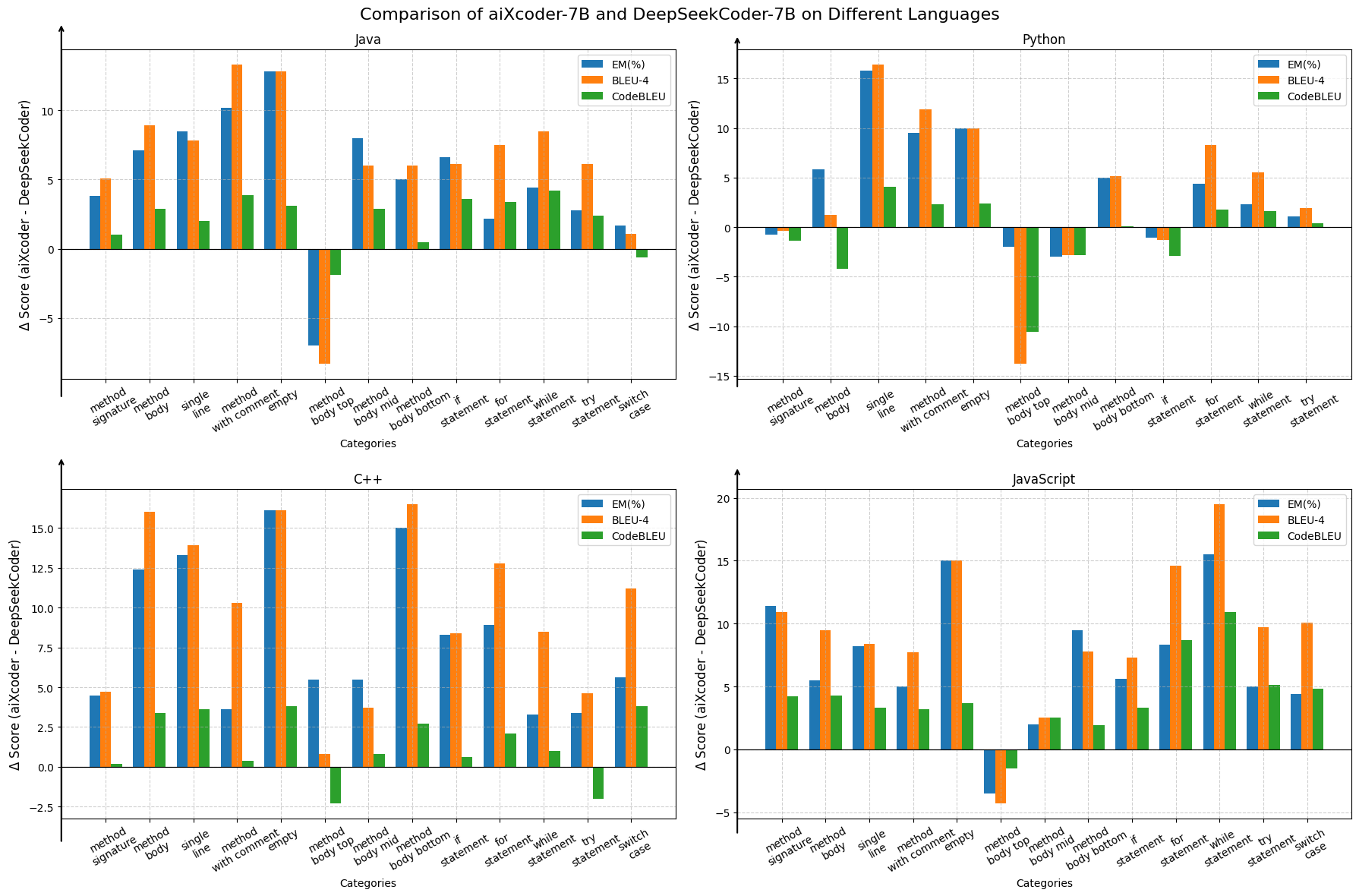}
\caption{Performance of LLMs on different types of code in FIM-Eval.}
\label{fig:cp}
\end{figure*}

Table \ref{tab:extent_data} shows the average generation performance on \textbf{FIM-Eval}. Figure \ref{fig:cp} shows the performance of LLMs in predicting different types of code. Based on the results, we obtain the following observations:
\begin{itemize}[leftmargin=*]
    \item In real-world programming, \model performs well in FIM. When evaluated on Java, C++, and JavaScript in FIM-Eval, aiXcoder-7B surpasses DeepSeekCoder-7B by an average of 5.2, 6.7, and 6.4 in FIM metrics for these three languages, highlighting its multilingual versatility. It is highest in C++, exceeding StarCoder2-7B by 11.8.
    \item \model offers no clear edge over DeepSeekCoder-7B in Python, likely due to lower training data proportion. When calculating CodeBLEU in FIM-Eval, \model's score of 63.0 is slightly lower than DeepSeekCoder-7B's score of 63.4. In several aspects, such as method body top/mid and if statement, it falls behind by up to 10\%, indicating the need for a better understanding of method initiations and conditional branches. Moreover, in the SantaCoder-data benchmark, \model's 83.0\% EM of Java is 5.1\% lower than the best score of 88.1\%. This will be rectified by boosting Python and Java data in training.
\end{itemize}

\subsection{RQ3: Performance on Cross-File Code Completion}
\label{sec:results:cross-file}

Another important capability of LLMs is the ability to understand code context across files, as developers often need to consider information from other files within the current project when writing code \cite{DevEval,EvoCodeBench}. In Table \ref{tab:CrossCodeEval}, we fix the context length for all LLMs at 16K and format the input using the PSM pattern in FIM. All LLMs employ the greedy search to generate code.

We design three experimental settings: 
\ding{182} \textbf{Base Setting.} As a baseline, LLMs predict the code based solely on the current file without cross-fire context.
\ding{183} \textbf{Retrieval BM25.} Based on the current file context in the base settings, it additionally uses BM25 to match repository code fragments. The top 5 matches, capped at 512 tokens, are added to the prompt.
\ding{184} \textbf{Retrieval w/Ref.} In this setting, we use not only the in-file context (as in Retrieval BM25 setting) but also the reference to retrieve the cross-file context. We prepend the retrieved context to the in-file context to construct the prompt for this setting.

The subsequent conclusions can be made:

\begin{itemize}[leftmargin=*]
    
    \item Under three experimental settings, \model performs very well, achieving EM of 30.0 and ES of 70.8 in Python, outperforming CodeLlama-7B by 7.7 and 21.4 in the base setting. In the other two experimental settings with retrieval, \model has an average EM that is higher than DeepSeekCoder-7B by 3.6 and 4.5 and an average ES that is higher by 4.5 and 8.6.
    \item The model's performance varies across different languages. \model excels in C\# and achieves an EM of 61 in the third setting. In other languages, the improvements of \model slightly decrease. For example, it only achieves an EM of 40.4 in Python. In the future, we will continue to enhance the model's performance across different languages.
\end{itemize}

%% file: table/santacoder_overall.tex
\begin{table}[ht]
    \caption{The exact match of LLMs on the SantaCoder-data benchmark.}
    \begin{small}
	\centering
	\resizebox{\linewidth}{!}{\begin{tabular}{lcccc}
			\toprule
			\multicolumn{1}{c}{Model}&Python&JavaScript&Java&Avg \\
               \midrule
                StarCoder2-7B&61.1\%&77.5\%&81.1\%&73.2\% \\
                CodeLlama-7B&67.6\%&74.3\%&80.2\%&74.0\% \\
                CodeLlama-13B&68.3\%&77.6\%&80.7\%&75.5\% \\
                DeepSeekCoder-7B&66.6\%&79.7\%&\textbf{88.1}\%&78.1\% \\
                \rowcolor[rgb]{ .741,  .843,  .933}
                \model &\textbf{73.3}\%&\textbf{81.7}\%&83.0\%&\textbf{79.3}\% \\
               \bottomrule
            \end{tabular}}
        \label{table:result-santacoder}
        \end{small}
\end{table}

%% file: table/code_gen_cross.tex
\begin{table*}[!htbp]
\centering
\caption{Performance of LLMs on CrossCodeEval.}
\resizebox{0.9\textwidth}{!}{
\begin{tabular}{l c c c c c c c c c c}
\toprule
\multirow{3}{*}{\bf Model} & \multicolumn{2}{c}{\textbf{Python }} & \multicolumn{2}{c}{\textbf{Java }} & \multicolumn{2}{c}{\textbf{TypeScript }} & \multicolumn{2}{c}{\textbf{C\# }} & \multicolumn{2}{c}{\textbf{Average}}\\
\cmidrule(lr){2-3} \cmidrule(lr){4-5} \cmidrule(lr){6-7} \cmidrule(lr){8-9} \cmidrule(lr){10-11}
& {EM} & {ES} & {EM} & {ES} & {EM} & {ES} & {EM} & {ES} & {EM} & {ES} \\\midrule
\multicolumn{11}{c}{\textbf{Base Model}} \\\midrule
CodeLlama-7B
& 22.3 & 55.2 & 27.9 & 66.9 & 10.8 & 70.9 & 45.8 & 77.2 & 26.7 & 67.6
\\
StarCoder2-7B
& 22.5 & 57.3 & 25.9 & 65.9 & 28.9 & 71.6 & 39.5 & 70.5 & 29.2 & 66.3
\\
DeepSeekCoder-7B
& 27.2 & 62.3 & 33.4 & 73.2 & 36.6 & 77.3 & 45.9 & 77.0 & 35.8 & 72.4
\\
\rowcolor[rgb]{ .741,  .843,  .933}
\model 
& \textbf{30.0} & \textbf{70.8} & \textbf{34.9} & \textbf{77.8} & \textbf{35.3} & \textbf{79.6} & \textbf{49.9} & \textbf{86.4} & \textbf{37.5} & \textbf{78.7}
\\\midrule
\multicolumn{11}{c}{\textbf{+ Retrieval BM25}} \\\midrule
CodeLlama-7B
& 23.5 & 53.5 & 33.9 & 68.4 & 11.5 & 71.5 & 50.6 & 75.3 & 29.9 & 67.2
\\
StarCoder2-7B
& 25.3 & 58.0 & 31.4 & 67.4 & 33.3 & 73.2 & 43.5 & 69.8 & 33.4 & 67.1
\\
DeepSeekCoder-7B
& 29.9 & 62.9 & 39.8 & 74.8 & 39.0 & 77.0 & 52.2 & 78.1 & 40.2 & 73.2
\\
\rowcolor[rgb]{ .741,  .843,  .933}
\model 
& \textbf{35.3} & \textbf{74.3} & \textbf{42.2} & \textbf{80.4} & \textbf{39.9} & \textbf{81.3} & \textbf{57.7} & \textbf{88.8} & \textbf{43.8} & \textbf{81.2}
\\
\midrule
\multicolumn{11}{c}{\textbf{+ Retrieval w/ Ref.}} \\\midrule
CodeLlama-7B
& 26.7 & 54.9 & 36.3 & 69.0 & 12.8 & 72.9 & 52.8 & 75.0 & 32.1 & 67.9
\\
StarCoder2-7B
& 28.5 & 59.0 & 35.0 & 69.2 & 36.0 & 72.6 & 47.9 & 71.6 & 36.9 & 68.1
\\
DeepSeekCoder-7B
& 33.2 & 64.5 & 43.7 & 76.1 & 43.4 & 78.4 & 55.4 & 78.7 & 43.9 & 74.4
\\
\rowcolor[rgb]{ .741,  .843,  .933}
\model 
& \textbf{40.4} & \textbf{76.3} & \textbf{47.0} & \textbf{82.4} & \textbf{45.0} & \textbf{83.8} & \textbf{61.0} & \textbf{89.4} & \textbf{48.4} & \textbf{83.0} \\
\bottomrule
\end{tabular}}
\label{tab:CrossCodeEval}
\end{table*}

%% file: table/extend_data.tex
\begin{table}[htbp]
\setlength{\tabcolsep}{3pt}
\centering
\caption{The performance of LLMs on FIM-Eval.}
\resizebox{0.85\linewidth}{!}{
\begin{tabular}{lcccc}
\midrule 
 \multirow{2}{*}{\textbf{Model}} & \multicolumn{4}{c}{\textbf{Java}} \\  \cmidrule{2-5}
 & \textbf{EM}(\%) & \textbf{BLEU-4} & \textbf{CodeBLEU} & \textbf{Average} \\
\midrule 
CodeLLama-7B & 38.1 & 56.9 & 69.9 & 55.0 \\
StarCoder2-7B & 37.7 & 57.7 & 69.2 & 54.9 \\
DeepSeekCoder-7B & 43.4 & 63.4 & 71.7 & 59.5 \\
Qwen2.5-Coder-7B & 46.0 & 65.3 & 71.6 & 60.9 \\
\rowcolor[rgb]{ .741,  .843,  .933}
\model & \textbf{49.4} & \textbf{70.6} & \textbf{74.0} & \textbf{64.7} \\
\midrule
\multirow{2}{*}{\textbf{Model}} & \multicolumn{4}{c}{\textbf{C++}} \\  \cmidrule{2-5}
 & \textbf{EM}(\%) & \textbf{BLEU-4} & \textbf{CodeBLEU} & \textbf{Average} \\
\midrule 
CodeLLama-7B & 25.4 & 44.2 & 63.9 & 44.5 \\
StarCoder2-7B & 22.6 & 41.9 & 61.2 & 41.9 \\
DeepSeekCoder-7B & 29.1 & 50.0 & 65.8 & 48.3 \\
Qwen2.5-Coder-7B & 30.3 & 48.6 & 62.4 & 47.1 \\
\rowcolor[rgb]{ .741,  .843,  .933}
\model & \textbf{37.3} & \textbf{60.3} & \textbf{67.4} & \textbf{55.0} \\
\midrule
\multirow{2}{*}{\textbf{Model}} & \multicolumn{4}{c}{\textbf{JavaScript}} \\  \cmidrule{2-5}
 & \textbf{EM}(\%) & \textbf{BLEU-4} & \textbf{CodeBLEU} & \textbf{Average} \\
\midrule 
CodeLLama-7B & 25.7 & 44.1 & 60.4 & 43.4 \\
StarCoder2-7B & 23.9 & 42.0 & 57.4 & 41.1 \\
DeepSeekCoder-7B & 29.3 & 49.0 & 60.3 & 46.2 \\
Qwen2.5-Coder-7B & 30.8 & 49.7 & 60.5 & 47.0 \\
\rowcolor[rgb]{ .741,  .843,  .933}
\model & \textbf{36.5} & \textbf{58.0} & \textbf{64.3} & \textbf{52.9} \\
\midrule
\multirow{2}{*}{\textbf{Model}} & \multicolumn{4}{c}{\textbf{Python}} \\  \cmidrule{2-5}
 & \textbf{EM}(\%) & \textbf{BLEU-4} & \textbf{CodeBLEU} & \textbf{Average} \\
\midrule 
CodeLLama-7B & 21.6 & 39.9 & 60.0 & 40.5 \\
StarCoder2-7B & 24.4 & 43.5 & 59.5 & 42.5 \\
DeepSeekCoder-7B & 29.6 & 51.1 & \textbf{63.4} & 48.0 \\
Qwen2.5-Coder-7B & 31.8 & 52.8 & 63.3 & 49.3 \\
\rowcolor[rgb]{ .741,  .843,  .933}
\model & \textbf{35.0} & \textbf{56.1} & 63.0 & \textbf{51.4} \\
\bottomrule
\end{tabular}}
\label{tab:extent_data}
\end{table}

%% file: chapter/7_Discussion.tex
\section{Discussion}
\label{sec:discussion}

\subsection{Why choose 7B?}
\label{sec:discussion:efficiency}

\input{table/efficiency}

The inference efficiency of code completion models is crucial for enhancing the developer’s experience. While scaling up model size can improve accuracy, it often comes at the cost of inference efficiency. Table \ref{tab:efficiency} illustrates the inference efficiency of our \model (bfloat16) and Qwen2.5-Coder-32B (int4) under identical testing settings on a single A100 GPU. Despite the quantization of Qwen2.5-Coder-32B to int4, which slightly reduces its inference quality, its inference speed and throughput are still significantly lower compared to \model in bfloat16. For example, Qwen2.5-Coder-32B requires 91.65 ms to generate a completion suggestion, whereas \model achieves a much faster response time of 29.26 ms—approximately 3 times faster. These results demonstrate that \model can generate completion suggestions both accurately and efficiently, making it superior for practical applications.

\subsection{The Comparison in the Length of Code}
\label{sec:discussion:length_comparison}

We propose a novel evaluation perspective in FIM, \ie comparing the code length between human-written reference code and code generated by LLMs. It is essential not only to ensure that the completed code is functionally correct but also that its length is consistent with what a human programmer would produce. 

To gain insights into this aspect, we evaluate LLMs' performance using FIM-Eval (Section \ref{sec:study_design:benchmarks}), which includes a variety of scenarios. Additionally, we present the code length ratio, which is calculated as the ratio of the number of tokens in the prediction to the number of tokens in the ground truth code. Based on the experimental results in Table \ref{tab:extent_ratios} below, we observe that existing LLMs tend to over-generate, producing code that is substantially longer than necessary. Too long code will increase the burden on users and reduce maintainability. For example, CodeLlama-7B produces ratios of 2.14 for Java and 3.02 for C++, while StarCoder2-7B reaches even higher ratios, such as 3.62 for C++. In contrast, \model consistently generates code predictions that are similar in length to human-written code, achieving ratios of 0.97 for Java and 0.87 for Python. We attribute this performance to aiXcoder-7B's Structured Fill-In-the-Middle (SFIM) training objective, which helps align model outputs with human-written code, resulting in more efficient coding practices.

\begin{table}[h]
\setlength{\tabcolsep}{3pt}
\centering
\caption{The length comparison between the generated code and reference code}
\vspace{-0.2cm}
\resizebox{\linewidth}{!}{
\begin{tabular}{lcccc}
\toprule
\textbf{Model} & \textbf{Java} & \textbf{C++} & \textbf{JavaScript} & \textbf{Python} \\
\midrule
CodeLlama-7B & 2.14(340/159) & 3.02(486/161) & 2.39(407/170) & 3.28(547/167) \\
StarCoder2-7B & 2.22(353/159) & 3.62(583/161) & 2.69(458/170) & 2.92(488/167) \\
DeepSeekCoder-7B & 1.37(217/159) & 2.05(330/161) & 1.37(232/170) & 1.65(275/167) \\
Qwen2.5-Coder-7B & 2.23(355/159) & 2.29(368/161) & 1.59(270/170) & 1.91(325/170) \\
\rowcolor[rgb]{ .741,  .843,  .933}
\model & 0.97(154/159) & 1.35(217/161) & 1.04(177/170) & 0.87(146/167) \\
\bottomrule
\end{tabular}}
\label{tab:extent_ratios}
\vspace{-0.4cm}
\end{table}

\subsection{Insights of Training LLMs for Code}
\label{sec:discussion:insights}

Based on our practices in \model, we summarize the following insights to help practitioners train the next generations of LLMs for code.

\textbf{Scaling up training data can continuously improve the performance of LLMs.} Although the scaling law \cite{scaling_law} provides a relationship between model size and the amount of training data, we discover that further scaling up training data is necessary. Even if the training loss of LLMs is already small, continually training with new data can still improve the performance of models. Similar phenomena have also been found in other works \cite{Grokking}.

\textbf{Exploiting the relationships between code files during training can enhance the LLMs' ability to understand cross-file context.} In practical applications, LLMs often need to predict code based on cross-file context \cite{CrossCodeEval}. Thus, during the training process, we should organize files based on their relationships and train LLMs to understand and utilize the cross-file context. For example, we sample files based on the similarity of their content, which is closer to retrieval-augmented code completion scenarios.

\textbf{Incorporating the code structures into the training objectives can improve the performance of LLMs.} 
The code is highly structured. However, previous LLMs \cite{DeepSeek-Coder,CodeLlama,StarCoder} view the code into plain text, ignoring the underlying code structures. This paper first incorporates code structures into the training objectives of LLMs and proposes SFIM. SFIM constructs code spans based on syntax trees of code and trains LLMs to generate accurate and concise code. The results in Section \ref{sec:results} show the effectiveness of our SFIM. This inspires practitioners to explore new training objectives to model valuable code structures, \eg control flows and data flows.

\subsection{Threats to Validity}
\label{sec:discussion:threats}

We summarize two main threats to this paper.

\textbf{Data Leakage.} A critical threat when training LLMs is the potential inclusion of evaluation data within the training set, which can undermine the reliability of evaluation outcomes. To address this threat, we exclude any data related to our evaluation datasets during data collection. Additionally, the FIM-Eval dataset we constructed and used in our experiments was further emphasized to ensure its independence from the training data. While we cannot guarantee the absence of data leakage in other models due to lack of access, our benchmarks demonstrate that \model outperforms them reliably.

\textbf{The selection of hyper-parameters.} Another threat to the validity of our study lies in the selection of hyperparameters and rules used during the training of aiXcoder-7B, including model architecture hyperparameters, thresholds for data cleaning, data deduplication parameters, code quality assessment criteria, and sensitive information removal strategies. We selected these based on our preliminary experiments and prior empirical knowledge. We did not conduct a comprehensive hyperparameter search due to the substantial computational costs involved in pre-training, potentially resulting in suboptimal configurations. However, this does not affect our contributions, as future improvements in hyperparameter optimization or heuristic rules can be easily integrated into our framework.

%% file: table/efficiency.tex
\begin{table}[h]
\setlength{\tabcolsep}{3pt}
\centering
\caption{The inference efficiency (\ie time taken to generate a completion suggestion) of \model (bfloat16) and Qwen2.5-Coder-32B (int4).}
\resizebox{\linewidth}{!}{
\begin{tabular}{cccc}
\toprule
\textbf{Input Length} & \textbf{Output Length} & \textbf{Qwen2.5-Coder-32B} & \textbf{\model} \\
\midrule
1024 & 128 & 23.70 ms & 14.81 ms \\
4096 & 128 & 32.71 ms & 17.10 ms \\
8192 & 128 & 64.70 ms & 21.79 ms \\
16384 & 128 & 91.65 ms & 29.26 ms \\
\bottomrule
\end{tabular}}
\label{tab:efficiency}
\end{table}

%% file: chapter/8_Related_Work.tex
\section{Related Work}
\label{sec:related_work}

This section provides an overview of the evolution of LLMs for code. We categorize LLMs into closed-source and open-source models.

\textbf{Closed-Source LLMs for Code.}
One of the earliest notable breakthroughs is Codex \cite{HumanEval}, introduced by OpenAI. Codex is fine-tuned from GPT-3 using a high-quality code dataset and demonstrates strong performance in Python code completion. 
Following Codex, OpenAI continued to lead with the development of GPT-4, GPT-4 Turbo, and GPT-4o, all of which exhibit strong capabilities in both code generation and code completion \cite{GPT4}. Alongside OpenAI, Google introduced Gemini \cite{team2023gemini}, and Anthropic contributed with Claude3 \cite{Claude3}, all of which integrate code-related tasks into broader conversational capabilities. These closed-source models dominate the landscape with outstanding performance in code-related tasks due to their extensive training datasets and advanced optimization techniques.

\textbf{Open-Source LLMs for Code.}
Parallel to the development of closed-source LLMs, open-source LLMs have significantly expanded access to code-related technologies, thereby fostering exploration and innovation in the research community. Meta AI's CodeLlama \cite{CodeLlama}, built upon Llama2 \cite{Llama2}, showcased advanced capabilities, including fill-in-the-blank and zero-shot programming. DeepSeek AI's DeepSeek Coder \cite{DeepSeek-Coder}, trained on a diverse dataset of 2 trillion tokens, offered multiple model sizes to meet various needs. The BigCode community released StarCoder \cite{StarCoder}, trained on The Stack v2 dataset \cite{TheStackDatasets}, which outperformed other open-source models in Python and other programming languages at the time. These models, along with newer iterations like DeepSeek Coder V2\cite{zhu2024deepseek}, have effectively reduced the performance gap with closed-source models while promoting transparency, reproducibility, and community-driven development.

Our \model is part of the open-source community's ongoing efforts to advance code completion. It outperforms existing LLMs with similar sizes in six code completion benchmarks, serving as a lightweight and effective function model for academia and industry. 

%% file: chapter/9_Conclusion.tex
\section{Conclusion and Future Work}
\label{sec:conclusion}

\textbf{Conclusion.} This paper presents \model, a lightweight and effective LLM for code completion. \model is trained with 1.2 trillion unique tokens and employs some novel training techniques, including diverse data sampling strategies and multi-objective training. We conduct extensive experiments on six code completion benchmarks covering six programming languages. The results show that \model outperforms the
latest six LLMs with similar sizes and even surpasses four larger LLMs (\eg CodeLLaMa-34B). We also provide some valuable insights for helping practitioners train the next generations of LLMs for code.

\textbf{Future Work.} In the future, we will train more powerful lightweight LLMs for code completion. Specifically, we plan to design a model architecture dedicated to code. Compared with Transformer, it can explicitly model code structures, \eg syntax structures. In addition, we will soon release instruction fine-tuned versions of \model to support more software engineering tasks, \eg code summarization and code repair.